\documentclass[conference]{IEEEtran}
\IEEEoverridecommandlockouts
\usepackage{cite}
\usepackage{amsmath,amssymb,amsfonts}
\usepackage{algorithmic}
\usepackage{graphicx}
\usepackage{textcomp}
\usepackage{xcolor}
\usepackage{hyperref}
\usepackage{booktabs}
\usepackage{multirow}
\usepackage{tabularx}
\usepackage[table]{xcolor}
\usepackage{pifont}

\def\BibTeX{{\rm B\kern-.05em{\sc i\kern-.025em b}\kern-.08em
    T\kern-.1667em\lower.7ex\hbox{E}\kern-.125emX}}
\usepackage{cuted}   
\usepackage{capt-of} 
\usepackage[markup=underlined]{changes}
\usepackage{svg}

\begin{document}
 
\title{LSA: Localized Semantic Alignment for Enhancing Temporal Consistency in Traffic Video Generation

\thanks{
\textsuperscript{1} Mercedes-Benz AG, Germany

\textsuperscript{2} ETH Zurich, Switzerland

\textsuperscript{3} Friedrich-Alexander University Erlangen-Nuremberg, Germany

\textsuperscript{*} Equal Contribution

\textsuperscript{\dag} Correspondence to mirlan.karimov@mercedes-benz.com

This work is funded by the Federal Ministry for Economic Affairs and Energy within the project "nxtAIM".}

} 


\author{%
\centering
\begin{tabular}{ccc}
\parbox[t]{0.32\textwidth}{\centering Mirlan Karimov\textsuperscript{1,2,*,\dag}} &
\parbox[t]{0.32\textwidth}{\centering Teodora Spasojević\textsuperscript{1,3,*}} &
\parbox[t]{0.32\textwidth}{\centering Markus Braun\textsuperscript{1}} \\[0.6em]
\parbox[t]{0.32\textwidth}{\centering Julian Wiederer\textsuperscript{1,3}} &
\parbox[t]{0.32\textwidth}{\centering Vasileios Belagiannis\textsuperscript{3}} &
\parbox[t]{0.32\textwidth}{\centering Marc Pollefeys\textsuperscript{2}} \\
\end{tabular}
}

\IEEEaftertitletext{%
\vspace{-2.0\baselineskip} 
\begin{center}
  \includegraphics[width=0.98\textwidth]{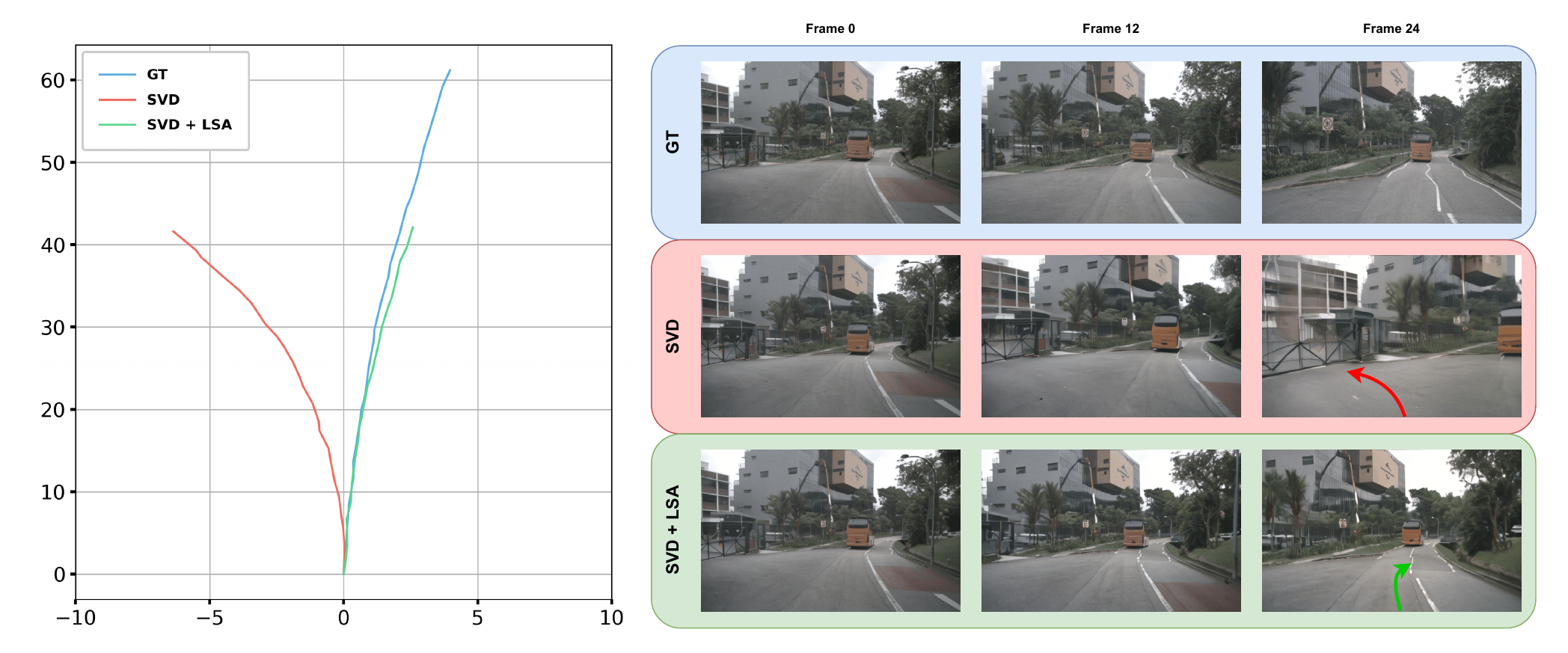}
  \captionof{figure}{Improved ego-motion by fine-tuning SVD \cite{svd} with our LSA framework. \textbf{Left}: Ego trajectories estimated by VGGT \cite{vggt} from the generated videos show that semantic feature alignment via LSA framework greatly improves temporal consistency and yields more accurate ego-motion that closely follows the ground truth trajectory. \textbf{Right}: The improvement is also evident in the generated videos from nuScenes \cite{nuscenes}, where the SVD that is fine-tuned without LSA exhibits drift into an unnatural path and degraded frame quality over time.}
  \label{fig:teaser}
\end{center}
}

\maketitle

\vspace{1cm}
\begin{abstract}
Controllable video generation has emerged as a versatile tool for autonomous driving, enabling realistic synthesis of traffic scenarios. However, existing methods depend on control signals at inference time to guide the generative model towards temporally consistent generation of dynamic objects, limiting their utility as scalable and generalizable data engines.
In this work, we propose Localized Semantic Alignment (LSA), a simple yet effective framework for fine-tuning pre-trained video generation models. LSA enhances temporal consistency by aligning semantic features between ground-truth and generated video clips. Specifically, we compare the output of an off-the-shelf feature extraction model between the ground-truth and generated video clips localized around dynamic objects inducing a semantic feature consistency loss. We fine-tune the base model by combining this loss with the standard diffusion loss. The model fine-tuned for a single epoch with our novel loss outperforms the baselines in common video generation evaluation metrics. To further test the temporal consistency in generated videos we adapt two additional metrics from object detection task, namely mAP and mIoU. Extensive experiments on nuScenes and KITTI datasets show the effectiveness of our approach in enhancing temporal consistency in video generation without the need for external control signals during inference and any computational overheads. The code is available at \url{https://github.com/mirlanium/LSA}

\end{abstract}



\section{Introduction}\label{sec:introduction}

Collecting real-world datasets for training and evaluating deep learning models in autonomous driving is inherently challenging. The safety-critical nature of autonomous systems demands coverage of rare long-tail events that are costly and often impractical to capture through real-world driving. Recent advances in video generation models have emerged as a scalable alternative, enabling the synthesis of diverse and realistic driving scenarios.

While diffusion-based video generation methods achieve high visual fidelity, modeling high-level world concepts such as object permanence and physically plausible motion directly in pixel space remains difficult due to the large sampling space \cite{e2echallenges}. As a result, generated sequences often exhibit temporal inconsistencies, including flickering, object deformation, and unrealistic motion trajectories of dynamic objects, which limit their utility in downstream perception and planning tasks. Therefore, such models offer limited value in the autonomous driving context, where the requirements on generated videos extend beyond visual realism.

Conditional video generation models specialized for driving scenarios \cite{panacea, drive_dreamer, magicdrivedit, ctrlv, magicdrive2, dive} have leveraged auxiliary conditioning signals to guide object placement and motion by conditioning the video generation model via control mechanisms \cite{controlnet, adaln, cross_attention}. While these control conditions act as strong anchors that improve temporal consistency, such methods rely on availability of control signals at inference time, which limits their scalability. Moreover, even under explicit conditioning, flickering and motion jitter remain persistent.

Recent efforts have sought to relax dependency on external control signals at inference time. Ctrl-V \cite{ctrlv} builds upon Stable Video Diffusion (SVD) \cite{svd} with a two-stage pipeline: the first stage infers bounding-box trajectories for dynamic objects, and the second stage generates video frames conditioned on the inferred boxes. The first stage of Ctrl-V comes in two variants. The first variant interpolates bounding boxes between the first and last frames, thus still requiring control signals at inference time. The second variant predicts forward from only the first-frame bounding box, hence relaxing the dependency, but leads to temporal inconsistencies. Moreover, Ctrl-V introduces additional model complexity due to its two-stage design and control mechanism.

Recent works on video diffusion models \cite{repa} find that representations learned by diffusion models are semantically weaker than those produced by DINOv2 \cite{dinov2}, a self-supervised image representation learning model. SG-I2V \cite{sg_i2v} shows that the feature maps extracted from the upsampling blocks of SVD \cite{svd} are only weakly aligned across frames.

Building upon these insights, we propose Localized Semantic Alignment (LSA), a simple yet effective framework for fine-tuning pre-trained video generation models to enhance temporal consistency without relying on external control inputs at inference time. We apply LSA on the publicly available SVD model [14]. LSA introduces a localized semantic alignment mechanism that integrates seamlessly into the SVD training pipeline. Specifically, we reconstruct generated video frames from denoised video latents and extract semantic feature embeddings using the DINOv2 model [16]. Then, a misalignment between semantic features of ground-truth and generated frames is computed, with greater emphasis on dynamic regions localized by ground-truth bounding boxes. We fine-tune SVD combining this loss and the standard diffusion loss. This fine-tuning framework encourages the model to learn consistent motion patterns and coherent object appearance, while keeping the inference pipeline identical to the original SVD. 

Our main contributions are:
\begin{itemize}
    \item We introduce LSA, a fine-tuning framework that improves the temporal consistency of pre-trained video generation models within a single epoch of fine-tuning.
    \item We demonstrate that LSA outperforms a two-step conditional video generation method in terms of both standard video quality and detection-based evaluation metrics, while being more computationally efficient.
    \item We show that SVD fine-tuned with LSA can serve as a drop-in enhancement for conditional video generation, consistently improving the quality and temporal coherence of generated videos.   
\end{itemize}

\section{Related Work}\label{sec:related_work}
This section reviews diffusion-based video generation models and their adaptation to autonomous driving. We further review recent methods proposing semantic feature alignment to improving temporal consistency.

\subsection{Video Generation Models}\label{subsec:videogen}
Diffusion-based video generation models have recently demonstrated strong capabilities in synthesizing high-fidelity videos. Early efforts in this domain extended image diffusion models \cite{ddpm} to the temporal domain using 3D U-Net \cite{3dunet} architecture enabling joint modeling of spatial and temporal correlations \cite{vdm}. Latent Video Diffusion \cite{lvdm} and Stable Video Diffusion \cite{svd}, adapt latent diffusion frameworks \cite{ldm} to the video domain, enabling efficient image-to-video generation conditioned on the first frame while maintaining high visual fidelity. Despite their success, these methods often struggle to model long-range dependencies and object permanence, resulting in flickering or inconsistent object appearances across frames. 

\subsection{Video Generation Models in Autonomous Driving}\label{subsec:videogenad}
Autonomous driving represents a challenging testbed for video generation, requiring realistic scene dynamics and consistent object motion beyond fidelity of generated videos. Such concepts prove difficult to model in pixel-space due to high sampling space \cite{e2echallenges}. Several approaches have attempted to explicitly condition video generation models on a variety of conditioning signals such as BEV scene layouts \cite{panacea}, HD maps \cite{drive_dreamer}, object bounding boxes \cite{magicdrivedit, ctrlv} and ego trajectories \cite{orbis} to guide the generation. Multi-view generation models \cite{dive, magicdrive2} have introduced cross-view attention \cite{driving_diffusion} to ensure multi-view consistency. While these methods achieve improved scene realism, they assume the availability of such conditioning signals both at the training and inference time limiting their utility as a scalable data engine for real-world scenario generation. In contrast, our approach introduces semantic regularization only during training. By enforcing semantic alignment, we induce strong temporal and semantic consistency in video generation without adding inference-time complexity and without relying on an external conditioning signal.

\subsection{Semantic Feature Alignment}\label{subsec:semanticfeat}
Recent advances in self-supervised representation learning \cite{dinov2, dino}, have made it possible to extract semantically rich and robust features without explicit supervision. Several works have leveraged such pre-trained features to improve semantic representations learned by diffusion models. REPA \cite{repa} improves both training efficiency and generation quality by aligning early diffusion-transformer hidden states to clean DINOv2 representations. NormalCrafter \cite{normal_crafter} aligns the diffusion features with robust semantic representations from DINO encoder for accurate and detailed normal estimation. Recent video generation models in autonomous driving have also demonstrated effectiveness of feature alignment with pre-trained features to enhance generation quality. For instance, the Orbis \cite{orbis} world model aligns the tokenizer’s semantic branch with DINOv2 features via cosine similarity during training while GAIA-1 \cite{gaia1} increases semantic content of image tokens predicted by the world model through DINO \cite{dino} distillation.

Closely related to our approach, SG-I2V \cite{sg_i2v} find that feature maps extracted from the upsampling blocks of video diffusion models are only weakly aligned across frames and align them by replacing the key and value tokens for each frame with those of the first frame. In contrast, to better handle the large frame-to-frame discrepancies in autonomous driving scenes, we align features of each generated frame with those of its corresponding ground-truth frame rather than enforcing consistency with the first frame.

\begin{figure*}[!t]
    \centering
    \includegraphics[width=\textwidth]{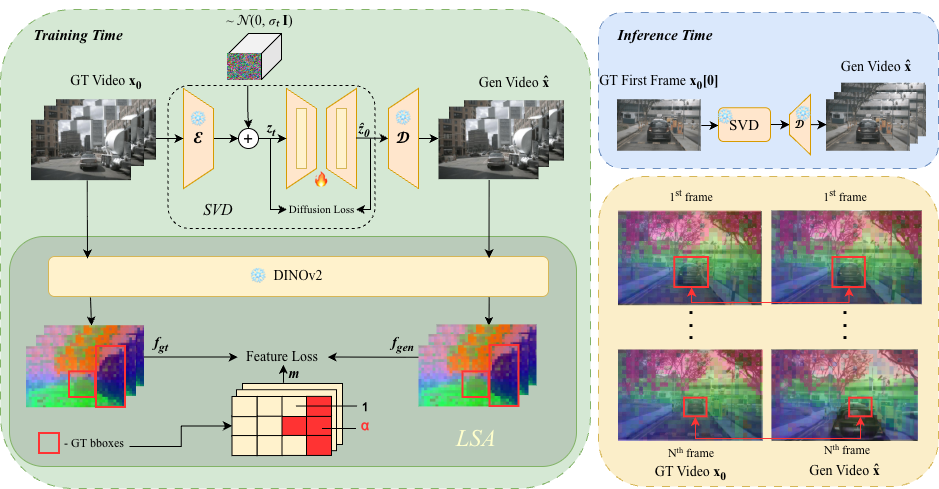}
    \caption{Overview of LSA, our proposed framework for improving temporal consistency in video generation. LSA introduces a semantic feature consistency loss that enforces alignment between the semantic representations of SVD-generated frames $\hat{\mathbf{x}}$ and their corresponding ground-truth frames $\mathbf{x}_{0}$, specifically within dynamic-object regions defined by ground-truth bounding boxes $\mathbf{bb}_{\text{gt}}$, promoting appearance consistency and temporally stable localization. Semantic features are extracted with DINOv2 \cite{dinov2}, while ground-truth bounding boxes provide spatial supervision. Inference stage of SVD fine-tuned with LSA is identical to the original SVD, hence not requiring bounding boxes. Example input frames are from nuScenes \cite{nuscenes}.}
    \label{fig:pipeline}
\end{figure*}

\section{Methodology}\label{sec:methodology}

In this section, we describe LSA, the proposed fine-tuning framework for enhancing the temporal consistency of video generation models. We apply LSA on the publicly available diffusion-based image-to-video generation model, SVD \cite{svd}. The framework introduces a localized semantic alignment mechanism that integrates seamlessly into the SVD training pipeline. Specifically, our method extracts semantically rich DINOv2 \cite{dinov2} features from reconstructed and ground-truth frames (Sec. \ref{subsec:feat_extraction}), computes localized feature misalignment around dynamic objects (Sec. \ref{subsec:feature_loss}), and combines this signal with the standard diffusion loss for multi-task fine-tuning. An overview of the complete framework is shown in Fig. \ref{fig:pipeline}.





\subsection{Video Reconstruction and Semantic Embedding Extraction}\label{subsec:feat_extraction}

\noindent\textbf{Preliminaries.} SVD \cite{svd} is a diffusion-based image-to-video generation model that operates in the latent space. It employs a Variational Autoencoder (VAE) \cite{vae}, consisting of encoder $\mathcal{E}$ and decoder $\mathcal{D}$, to compress the input ground-truth video $\mathbf{x}_0 \in \mathbb{R}^{N \times C \times H \times W}$ into latent representations $\mathbf{z}_0 \in \mathbb{R}^{N \times c \times h \times w}$, where $N$ is the number of frames, $C$ is the number of channels in the input video, $c$ the number of channels in the latent representation, $H \times W$ the input frame resolution, and $h \times w$ the latent spatial dimensions. A noise scheduler \cite{euler} is applied to the encoded video frames, producing noisy latents $\mathbf{z}_t$, where $t$ denotes the number of diffusion steps. In parallel, the first ground-truth frame is encoded by the encoder $\mathcal{E}$, replicated $N$ times, and concatenated with the noisy latents $\mathbf{z}_t$ along the channel dimension, allowing it to guide the denoising process toward the initial frame’s semantics. The concatenated latents are processed by the SVD's U-Net \cite{unet} backbone over $t$ diffusion steps.
Additionally, the first ground-truth frame is encoded by CLIP \cite{clip}, producing condition $\mathbf{c}_0$ and injected into the U-Net \cite{unet} backbone through cross-attention layers. 






SVD's U-Net produces a prediction of the latent velocity $\mathbf{v}_\theta(\mathbf{z}_t; \mathbf{c}_0, t)$. Following the \textit{v}-prediction parameterization \cite{v_pred, imagen_video}, this prediction is converted to a denoised latent estimate:
\begin{equation}
    \hat{\mathbf{z}}_0 = c_{\text{out}}(\sigma)\,\mathbf{v}_\theta(\mathbf{z}_t; \mathbf{c}_0, t) + c_{\text{skip}}(\sigma)\,\mathbf{z}_t,
\end{equation}
where $c_{\text{out}}(\sigma) = -\frac{\sigma}{\sqrt{1 + \sigma^2}}$ and $c_{\text{skip}}(\sigma) = \frac{1}{1 + \sigma^2}$ are scheduler-dependent coefficients and $\sigma$ denotes the noise level at step $t$. The model is trained to minimize a scheduler-weighted mean squared error (MSE) between the predicted and the ground-truth clean latents:
\begin{equation}
    \mathcal{L}_{\text{diff}} = 
    \mathbb{E}_{\mathbf{z}_0, t} \!\left[
        w(\sigma)\,
        \big\|
            \hat{\mathbf{z}}_0 - \mathbf{z}_0
        \big\|_2^2
    \right],
    \quad
    w(\sigma) = \frac{1 + \sigma^2}{\sigma^2}.
    \label{eq:diffusion_loss}
\end{equation}
This diffusion loss serves as the sole training objective in SVD. Fig. \ref{fig:pipeline} visualizes SVD's training set-up, consisting of the VAE's encoder $\mathcal{E}$ \cite{vae} and U-Net \cite{unet} backbone.



\vspace{1em}

\noindent\textbf{Video reconstruction.} Our LSA method first adds the pre-trained decoder $\mathcal{D}$ \cite{svd}, originally applied only in SVD's inference framework, to the training framework, this way enabling the computation of our frame-grounded loss. Following the completion of the $T$ diffusion steps by the U-Net backbone, the decoder reconstructs the final generated video frames $\mathbf{\hat{x}}$ from the denoised latent representation $\mathbf{\hat{z}}_0$, as illustrated in Fig. \ref{fig:pipeline}.


\vspace{1em}

\noindent\textbf{Semantic feature extraction.} Next, LSA extracts semantically rich feature maps from both ground-truth $\textbf{x}_0$ and generated $\mathbf{\hat{x}}$ video frames, as illustrated in Fig \ref{fig:pipeline}. For feature extraction, we adopt the pre-trained, publicly available DINOv2 \cite{dinov2}, a self-supervised Vision Transformer (ViT) whose representations capture explicit semantic segmentation of objects present in the input image. Given an input image of size $H \times W$, DINOv2 partitions it into non-overlapping patches of size $p \times p$, resulting in a grid of $\frac{H}{p} \times \frac{W}{p}$ patches. Each patch is embedded into a feature vector of dimension $d$, yielding a feature map $\mathbf{f} \in \mathbb{R}^{\frac{H}{p} \times \frac{W}{p} \times d}$. These semantically meaningful representations provide a robust basis for measuring content alignment between generated and reference ground-truth videos. 

For the $i$-th video and $j$-th frame, the feature maps of the ground-truth and generated frames are obtained as:
\begin{equation}
    \mathbf{f}_{\text{gt}}^{(i,j)} = \text{DINOv2}(\mathbf{x}_0^{(i,j)}), \quad
    \mathbf{f}_{\text{gen}}^{(i,j)} = \text{DINOv2}(\mathbf{\hat{x}}^{(i,j)}).
\label{eq:feature_extraction}
\end{equation}

\subsection{Semantic Feature Consistency Loss}\label{subsec:feature_loss}

\noindent\textbf{Dynamic objects masking.} To specifically improve the generation of dynamic objects, we construct dynamic object masks for the per-frame feature grids from Eq. \ref{eq:feature_extraction}. In these masks, each feature vector corresponding to a patch that overlaps with a ground-truth bounding box of a dynamic object in the input video is assigned a higher weight $\alpha > 1$, while all other feature vectors are assigned a weight of $1$, as shown in Fig. \ref{fig:pipeline}. Formally, the mask for the $i$-th video and $j$-th frame is defined as:

\begin{equation}
    \mathbf{m}^{(i,j)}(u,v) =
    \begin{cases}
        \alpha, & \text{if } (u,v) \!\in\! \mathcal{P}_{\text{dyn}}^{(i,j)} ,\\
        1, & \text{otherwise.}
    \end{cases}
\label{eq:mask}
\end{equation}

\noindent where $(u,v)$ indexes the 2D patch location on the $\tfrac{H}{p}\!\times\!\tfrac{W}{p}$ feature grid (row $u\!\in\!\{1,\dots,\tfrac{H}{p}\}$, column $v\!\in\!\{1,\dots,\tfrac{W}{p}\}$) and $\mathcal{P}_{\text{dyn}}^{(i,j)}$ denotes the set of patches in frame $j$ of video $i$ that fall into spatial regions within the ground-truth bounding boxes $\mathbf{bb}_{\text{gt}}^{(i,j)}$ of dynamic objects.

By applying these masks to the feature loss, LSA prioritizes the semantic alignment of features corresponding to the ground-truth locations of dynamic objects while maintaining global consistency across all regions.

\vspace{1em}

\noindent\textbf{Semantic feature consistency loss.} Recall that our goal is to align the semantic features of the ground-truth input and generated video frames, $\mathbf{x}_0$ and $\mathbf{\hat{x}}$, respectively, \textit{with an emphasis on improving alignment in regions corresponding to the ground-truth bounding boxes of dynamic objects}, $\mathbf{bb}_{\text{gt}}$. To achieve this, we define the last building block of our LSA framework, the semantic feature consistency loss, an MSE loss between the feature maps of the ground-truth and generated frames, $\mathbf{f}_{\text{gt}}$ and $\mathbf{f}_{\text{gen}}$, weighted by the dynamic object mask $\mathbf{m}$:

\begin{equation}
    \mathcal{L}_{\text{feat}} = \mathbb{E}_{i,j} \Big[ \| (\mathbf{f}_{\text{gt}}^{(i,j)} - \mathbf{f}_{\text{gen}}^{(i,j)}) \odot \mathbf{m}^{(i,j)} \|_2^2  \Big],
\label{eq:feature_loss}
\end{equation}

\noindent where $i$ denotes the video batch index, and $j$ denotes the time step of the video, ranging from $1$ to $N$. This formulation enforces semantic feature alignment between corresponding frames while emphasizing regions containing dynamic objects through $\mathbf{m}$.

\vspace{1em}

Finally, the proposed semantic feature consistency loss is combined with the original diffusion loss to form the overall training objective. During training, only the U-Net backbone of SVD is kept trainable, while all other pre-trained components, including the VAE's encoder and decoder, and DINOv2, are kept frozen. As visualized in Fig. \ref{fig:pipeline}, in the inference framework, SVD is left untouched.

Fig. \ref{fig:pipeline} illustrates an example video for interpreting the proposed semantic feature consistency loss. In the final frames of the ground-truth and generated clips, $\mathbf{x}_0$ and $\mathbf{\hat{x}}$, clear discrepancies in vehicle dynamics can be observed. Features within the ground-truth bounding boxes differ notably in the generated frame, indicating inconsistent motion generation and resulting in a high loss value.

\newcolumntype{M}[1]{>{\centering\arraybackslash}m{#1}}
\newcolumntype{P}[1]{>{\centering\arraybackslash}p{#1}}
\newcolumntype{Y}{>{\centering\arraybackslash}X} 

\begin{table*}[t]
\centering
\caption{Comparison of visual fidelity, downstream detection, and inference efficiency across models and datasets. Best in \textbf{bold}, second best \underline{underlined}.}
\label{tab:main_results}
\footnotesize
\setlength{\tabcolsep}{2pt}
\renewcommand{\arraystretch}{1.05}

\begin{tabularx}{\textwidth}{@{}
  M{0.09\textwidth} |
  P{0.15\textwidth} |
  Y Y |
  Y Y |
  Y Y
@{}}
\toprule
\multicolumn{1}{c|}{\multirow{2}{*}{\textbf{Dataset}}} &
\multicolumn{1}{c|}{\multirow{2}{*}{\textbf{Model}}} &
\multicolumn{2}{c|}{\textbf{Visual quality}} &
\multicolumn{2}{c|}{\textbf{Downstream detection}} &
\multicolumn{2}{c}{\textbf{Inference efficiency}} \\
\cline{3-4}\cline{5-6}\cline{7-8}
& &
\textbf{FVD}$\downarrow$ &
\textbf{FID}$\downarrow$ &
\textbf{mAP}$\uparrow$ &
\textbf{mIoU}$\uparrow$ &
\textbf{Time (h)}$\downarrow$ &
\textbf{\#Params (M)}$\downarrow$ \\
\midrule
\multirow{4}{*}{nuScenes \cite{nuscenes}} &
  SVD \cite{svd} & 841.34 & 33.16 & 6.67 & 74.69 & 8 & 2254 \\
& SVD Fine-tuned \cite{svd} & \underline{256.12} & \underline{19.67} & 16.75 & 79.08 & 8  & 2254 (+\textcolor{blue}{0}\%) \\
& Ctrl-V 1-to-0 \cite{ctrlv} & 265.68 & 20.10 & \underline{18.14} & \underline{80.11} & 24 & 5189 (+\textcolor{red}{130}\%) \\
\rowcolor{gray!10}
& SVD + LSA (Ours) & \textbf{229.26} & \textbf{18.08} & \textbf{24.92} & \textbf{80.6} & 8 & 2254 (+\textcolor{blue}{0}\%) \\
\midrule
\multirow{4}{*}{KITTI \cite{kitti}} &
  SVD \cite{svd} & 1111.99 & 66.94 & 5.35 & 75.01 & 8  & 2254 \\
& SVD Fine-tuned \cite{svd} & 743.56 & 51.59 & 15.59 & 82.99 & 8  & 2254 (+\textcolor{blue}{0}\%) \\
& Ctrl-V 1-to-0 \cite{ctrlv} & \underline{669.08} & \underline{48.40} & \underline{18.08} & \underline{83.66} & 24 & 5189 (+\textcolor{red}{130}\%) \\
\rowcolor{gray!10}
& SVD + LSA (Ours) & \textbf{608.35} & \textbf{45.68} & \textbf{22.59} & \textbf{85.23} & 8  & 2254 (+\textcolor{blue}{0}\%) \\
\bottomrule
\end{tabularx}
\end{table*}

\section{Experiments}\label{sec:experiments}

In this section, we evaluate our proposed LSA framework for improving temporal consistency in image-to-video generation. We outline the experimental setup, present quantitative and qualitative comparisons with baselines, and conclude with ablation studies analyzing the design choices of our method.

\subsection{Experimental Setup}\label{subsec:setup}

\noindent\textbf{Implementation details.} In all of our experiments, we make use of the publicly available pre-trained image-to-video version of Stable Video Diffusion \cite{svd} and generate videos with 25 frames at a frame rate of 7 fps, and of spatial resolution $320 \times 512$. Our implementation builds on the SVD pipeline implementation provided by \cite{ctrlv}. Following the training and evaluation protocol of \cite{ctrlv} we adopt the default Discrete Euler scheduler \cite{euler} with $T = 50$ diffusion steps and AdamW optimizer \cite{adam} with learning rate $\eta = 1 \times 10^{-5}$. For feature extraction, we use DINOv2 (ViT-B/14), the \emph{base} model with patch size $p=14$ and feature vector dimension $d=768$. The dynamic object mask weight is determined empirically and set to $\alpha=10$ across all experiments.

\vspace{1em}

\noindent\textbf{Training and evaluation setup.} For the evaluation, following \cite{ctrlv}, we sample 200 non-overlapping test videos, and generate 25 frame long videos based on their first frames. The same conditional first frames are used in all experiments. We employ a two-stage training: (i) first, we fine-tune SVD for one epoch using only the diffusion loss; (ii) then, in the second epoch, we enable the proposed semantic feature consistency loss of our LSA framework with weight $\lambda_{\text{feat}}=100$ in the nuScenes, and $\lambda_{\text{feat}}=60$ in the KITTI case, while down-weighting the diffusion loss by a factor $0.9$. The weights are selected such that the magnitudes of both losses are on a comparable scale, and the final values are determined empirically from the set of candidates satisfying this criterion. The overall objective in the second epoch becomes $\mathcal{L}=0.9\,\mathcal{L}_{\text{diff}}+\lambda_{\text{feat}}\mathcal{L}_{\text{feat}}$. We fine-tune the U-Net backbone directly and do not employ LoRA adapters \cite{lora}. All of these design choices are ablated in Sec. \ref{subsec:abl_study}.

\vspace{1em}

\noindent\textbf{Baselines.} We first compare against vanilla SVD \cite{svd} and SVD \cite{svd} fine-tuned for two epochs on each dataset matching the total number of epochs of our two-stage fine-tuning setup. This model receives no conditional bounding box input and therefore serves as an uncontrolled reference.


We next consider Ctrl-V \cite{ctrlv}, which also builds on the same image-to-video SVD backbone but enforces consistency through explicit bounding-box conditioning at inference. Ctrl-V comprises (i) a Bbox Generator that, given ground-truth bounding boxes of the first one or three frames and the last frame, interpolates the intermediate box trajectories and (ii) a Box2Video module which is a fine-tuned SVD model conditioned via ControlNet \cite{controlnet} to follow these trajectories. We consider the Ctrl-V version \emph{Ctrl-V 1-to-0}, which omits the last-frame ground-truth boxes at inference and training, thereby requiring genuine motion and video prediction, rather than interpolation, making it more comparable to SVD in terms of usage of explicit conditioning signals. We implement and train the 1-to-0 variant, following their standard protocol, to ensure fair comparison.

In the Ctrl-V design, bounding boxes act as spatio-temporal anchors that constrain object placement during the inference. By contrast, our LSA method leverages ground-truth boxes only during training, leaving inference unchanged. This makes the comparison well-posed: all approaches share the same backbone, data and number of training epochs; fine-tuned SVD lacks an explicit consistency prior or conditioning, \emph{Ctrl-V 1-to-0} performs box prediction and imposes box-conditioned consistency at test time, while our LSA method learns a dynamics prior from the proposed loss, without requiring box conditions at inference.

\vspace{1em}

\noindent\textbf{Datasets and evaluation metrics.} Following common practice in autonomous driving video generation, we evaluate on nuScenes \cite{nuscenes} and KITTI Tracking \cite{kitti} datasets. From nuScenes, we use only the single-view front-facing RGB camera and obtain 2D bounding boxes by projecting the provided 3D annotations onto the image plane. For KITTI, which provides 2D and 3D boxes for the training split, we follow the protocol in \cite{ctrlv} and partition the training data into a training subset (scenes 1–19) and a test subset (scenes 20–21) to enable evaluation of Ctrl-V baseline that requires ground-truth boxes at both training and inference.

We report standard video and image fidelity metrics: Fréchet Video Distance (FVD) \cite{fvd} and Fréchet Inception Distance (FID) \cite{fid}. Beyond these standard metrics, recent works \cite{dreamforge, magicdrive, drivegen, drivewm} propose applying widely adopted object detection metrics, such as mean Average Precision (mAP) \cite{coco, pascal_voc} and mean Intersection over Unit (mIoU) \cite{pascal_voc}, to generated videos, thereby effectively evaluating object placement in generated videos.

For all reported metrics, we evaluate on 200 non-overlapping videos of 25 frames each. For mAP and mIoU, frames are kept at the original resolution of $320\times512$. For visual quality metrics, frames are resized to $256\times410$ following \cite{ctrlv}. Detections for mAP and mIoU are obtained with DN-DETR \cite{dn_detr} object detection model (applied identically to both ground-truth and generated frames), and we report metrics only for objects larger than $76\times76$ pixels and belonging to vehicle classes (car, motorcycle, bicycle, bus, train, truck), reflecting our training focus on improving the generation of relatively salient dynamic objects.

\subsection{Quantitative Results}

Table~\ref{tab:main_results} summarizes visual fidelity, downstream detection, and inference efficiency across models and datasets. Across both datasets, SVD fine-tuned in our LSA framework \emph{consistently outperforms} the SVD baseline and the \emph{Ctrl-V 1-to-0} variant, even though the latter uses $\times 2.3$ more parameters and $\times 3$ longer inference time. In contrast, our approach modifies \emph{only} the SVD training objective, while inference remains unchanged, leading to \emph{no} runtime or memory overhead and without any condition requirements. 


Quantitatively, we obtain consistent gains over fine-tuned SVD on both nuScenes and KITTI across all metrics. FVD is reduced by $10.5\%$ and $18.2\%$ and FID by $8.1\%$ and $6.6\%$ on nuScenes and KITTI, respectively. The downstream detection improves substantially: in mAP by $+48.8\%$ on nuScenes and $+44.9\%$ on KITTI, indicating better spatial grounding of dynamic objects. Against \emph{Ctrl-V 1-to-0}, SVD fine-tuned with our LSA method remains superior despite the baseline’s heavier inference: FVD improves by $15.8\%$ and $9.1\%$, FID by $10.0\%$ and $5.6\%$, and mAP by $37.4\%$ and $24.5\%$ on nuScenes and KITTI, respectively, confirming that a training-time, localized feature regularizer is more effective than explicit box prediction and conditioning at test time. mIoU is also consistently improved compared to both SVD (nuScenes: $+2.0\%$, KITTI: $+2.7\%$) and \emph{Ctrl-V 1-to-0} (nuScenes: $+0.7\%$, KITTI: $+6.4\%$), though the relative gain is smaller than for mAP. This behavior is expected, as mAP reflects detection improvements in \textit{classification confidence} as well as \textit{localization accuracy}, whereas mIoU is a region-overlap measure and thus less sensitive to subtle appearance and boundary refinements introduced by our loss.


\vspace{1em}

\begin{table}[!t]
\centering
\caption{Integrating Consistency-Aware SVD into Controllable Video Generation. Best in \textbf{bold}, second best \underline{underlined}.}
\label{tab:ctrlv_integration}
\scriptsize
\renewcommand{\arraystretch}{1.05}

\newcolumntype{S}{>{\centering\arraybackslash}p{0.10\columnwidth}}
\newcolumntype{M}{>{\centering\arraybackslash}p{0.19\columnwidth}}
\newcolumntype{L}{>{\centering\arraybackslash}p{0.26\columnwidth}}
\setlength{\tabcolsep}{2pt}

\begin{tabularx}{\columnwidth}{@{} S | M | L | c c c c @{}}
\toprule
\textbf{Dataset} & \textbf{Model} & \textbf{Video gen.\ backbone} & \textbf{FVD} $\downarrow$ & \textbf{FID} $\downarrow$ & \textbf{mAP} $\uparrow$ & \textbf{mIoU} $\uparrow$ \\
\midrule
\multirow{4}{*}{\parbox[c]{0.10\columnwidth}{\centering nuScenes \cite{nuscenes}}}
  & \multirow{2}{*}{Ctrl-V 1-to-1 \cite{ctrlv}} 
  & SVD fine-tuned \cite{svd} & \underline{224.44} & \underline{18.81} & \underline{38.59} & \textbf{85.51} \\
\multicolumn{1}{c|}{} & 
\multicolumn{1}{c|}{} & 
SVD + LSA (Ours)  & \textbf{199.37} & \textbf{17.29} & \textbf{43.93} & \underline{85.31} \\
\cmidrule(lr){2-7}
\multicolumn{1}{c|}{} & \multirow{2}{*}{Ctrl-V 3-to-1 \cite{ctrlv}} 
  & SVD fine-tuned \cite{svd} & \underline{209.26} & \underline{18.76} & \underline{47.69} & \underline{90.08} \\
\multicolumn{1}{c|}{} & 
\multicolumn{1}{c|}{} & 
SVD + LSA (Ours)  & \textbf{189.76} & \textbf{16.99} & \textbf{53.75} & \textbf{90.69} \\
\midrule
\multirow{4}{*}{\parbox[c]{0.10\columnwidth}{\centering KITTI \cite{kitti}}}
  & \multirow{2}{*}{Ctrl-V 1-to-1 \cite{ctrlv}} 
  & SVD fine-tuned \cite{svd} & \underline{653.84} & \underline{49.27} & \underline{22.35} & \underline{86.22} \\
\multicolumn{1}{c|}{} & 
\multicolumn{1}{c|}{} & 
SVD + LSA (Ours)  & \textbf{543.14} & \textbf{45.23} & \textbf{27.61} & \textbf{87.67} \\
\cmidrule(lr){2-7}
\multicolumn{1}{c|}{} & \multirow{2}{*}{Ctrl-V 3-to-1 \cite{ctrlv}} 
  & SVD fine-tuned \cite{svd} & \underline{644.72} & \underline{49.89} & \underline{27.58} & \underline{87.45} \\
\multicolumn{1}{c|}{} & 
\multicolumn{1}{c|}{} & 
SVD + LSA (Ours)  & \textbf{529.80} & \textbf{45.30} & \textbf{32.15} & \textbf{88.46} \\
\bottomrule
\end{tabularx}
\end{table}

\noindent\textbf{Integrating semantically aligned SVD into controllable video generation.} To further test the generality of our semantic feature consistency loss, we integrate SVD fine-tuned by our LSA framework into the controllable generation pipelines of \emph{Ctrl-V 1-to-1} and \emph{Ctrl-V 3-to-1}. These settings rely on bounding-box interpolation rather than prediction and represent conditional video generation rather than pure video prediction. As shown in Table \ref{tab:ctrlv_integration}, replacing the fine-tuned SVD backbone with our consistency-regularized version consistently improves both video-fidelity and detection metrics. This demonstrates that our loss enhances the underlying generative capacity of the model and can transfer effectively to controllable video generation frameworks.

\begin{figure}[!t]
  \centering
  \includegraphics[width=\columnwidth,keepaspectratio]{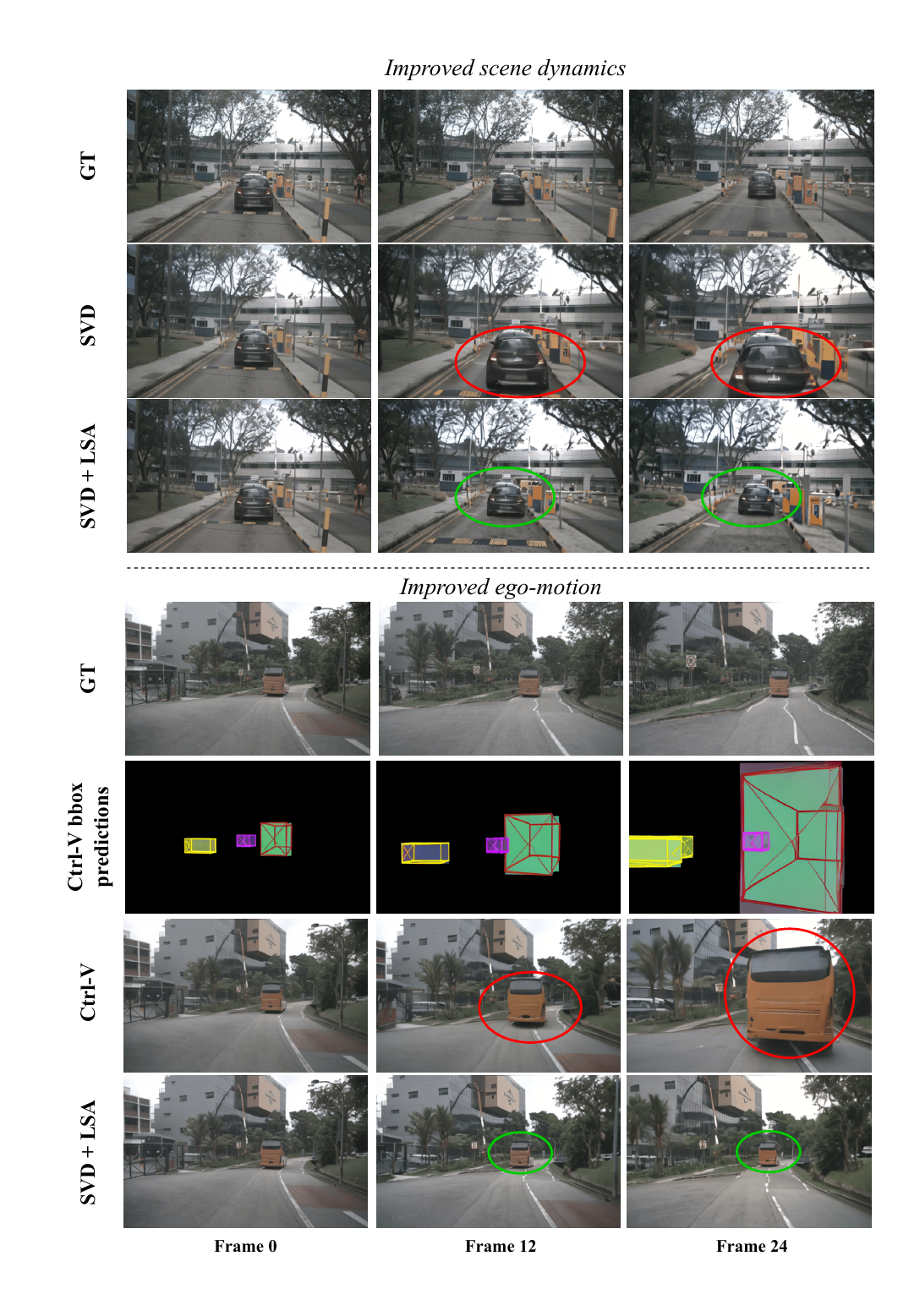}
  \caption{Visual comparison of our method with SVD \cite{svd} (top) and \textit{Ctrl-V 1-to-0} \cite{ctrlv} (bottom) on nuScenes \cite{nuscenes}. Our method yields more temporal consistency in motion dynamics of surrounding traffic agents (top) and ego-motion (bottom). \textcolor{red}{Red} circles mark inconsistent motions, while \textcolor{green}{green} circles mark correct ones.}
  \label{fig:qualitative_analysis_fig}
\end{figure}

\subsection{Qualitative Results}

Fig. \ref{fig:teaser} and Fig. \ref{fig:qualitative_analysis_fig} illustrate the improvements in temporal consistency achieved by our LSA fine-tuning framework. With just a single epoch of fine-tuning, our method produces more plausible ego-motion compared to SVD \cite{svd}, as visualized in Fig. \ref{fig:teaser} and estimated by the inverse dynamics model VGGT \cite{vggt}, which maps videos to pose sequences. LSA further enhances scene dynamics, as highlighted in Fig. \ref{fig:qualitative_analysis_fig} (top), leading to more stable, as well as visually richer scenes, with improved consistency in appearance of dynamic objects.

Furthermore, as shown in Fig. \ref{fig:qualitative_analysis_fig} (bottom), our approach outperforms \emph{Ctrl-V 1-to-0} \cite{ctrlv}, which suffers from inaccurate bounding-box predictions. These issues in Ctrl-V result in incorrect motion generation. In contrast, LSA mitigates these problems, producing consistent object motion across frames and maintaining coherent scene dynamics.

\subsection{Ablation Studies}\label{subsec:abl_study}

\noindent\textbf{Semantic feature consistency loss selection.}
In Table \ref{tab:feat_loss_types_abl_study} we analyze the effect of different formulations of our semantic feature consistency loss. We consider three variants, all instantiated from Eqs. \ref{eq:mask} and \ref{eq:feature_loss}: (i) \emph{Hybrid (Global + localized)}, which computes the loss over the entire feature grid while up-weighting patches that overlap dynamic objects $(\alpha>1$ inside boxes, 1 elsewhere) as described in Sec. \ref{sec:methodology}; (ii) \emph{Box-only}, which restricts feature alignment to dynamic-object regions by setting the mask to $\alpha$ for box-overlap and 0 otherwise; and (iii) \emph{Global-only (Uniform)}, which enforces global alignment without spatial emphasis by setting the mask uniformly to 1.

Empirically, \emph{Global-only} does not improve mAP or mIoU, indicating that uniform feature alignment is insufficient for learning correct object dynamics. In contrast, \emph{Box-only} yields gains in both metrics, suggesting that attending to dynamic-object regions leads to better dynamics understanding and improved object placement. The \emph{Hybrid} variant performs best as it couples strong localized supervision with global context.

\begin{table}[t]
\centering
\caption{Ablation of different semantic feature consistency losses on nuScenes dataset.}
\label{tab:feat_loss_types_abl_study}
\footnotesize
\setlength{\tabcolsep}{4pt}
\renewcommand{\arraystretch}{1.05}
\begin{tabular}{@{} l l c c c c @{}}
\toprule
\textbf{Model} & \textbf{Feature loss version} & \textbf{mAP} $\uparrow$ & \textbf{mIoU} $\uparrow$ \\
\midrule
SVD Fine-tuned \cite{svd} & N/A & 16.75 & 79.08 \\
SVD + LSA (Ours)        & Global-only (Uniform) & 14.22 & 75.54 \\
SVD + LSA (Ours)        & Box-only & \underline{17.97} & \underline{79.96} \\
\rowcolor{gray!10}
SVD + LSA (Ours)        & Hybrid (Global + localized) & \textbf{24.92} & \textbf{80.65} \\
\bottomrule
\end{tabular}
\end{table}

\vspace{1em}

\begin{table}[!t]
\centering
\caption{Comparison of different training strategies for our semantic feature consistency loss on nuScenes.}
\label{tab:unet_lora_abl_study}
\footnotesize
\setlength{\tabcolsep}{4pt}
\renewcommand{\arraystretch}{1.0}
\resizebox{\columnwidth}{!}{%
\begin{tabular}{@{} l l c c c c @{}}
\textbf{Model} & \textbf{Training strategy} & \textbf{FVD} $\downarrow$ & \textbf{FID} $\downarrow$ & \textbf{mAP} $\uparrow$ & \textbf{mIoU} $\uparrow$ \\
\midrule
SVD Fine-tuned \cite{svd} & N/A & \underline{256.12} & 19.67 & 16.75 & 79.08 \\
SVD + LSA (Ours)        & U-Net \cite{unet} (joint)  & 267.65 & 19.11 & 18.57 & 74.78 \\
SVD + LSA (Ours)        & LoRA \cite{lora} (staged) & 264.63 & \underline{18.73} & \underline{18.75} & \underline{80.11} \\
\rowcolor{gray!10}
SVD + LSA (Ours)        & U-Net \cite{unet} (staged)  & \textbf{229.26} & \textbf{18.08} & \textbf{24.92} & \textbf{80.65} \\
\bottomrule
\end{tabular}%
}
\end{table}

\noindent\textbf{Fine-tuning strategies comparison.} We evaluate two schedules for integrating our proposed loss: (i) \emph{joint} - fine-tune SVD from the first epoch with the combined objective $\mathcal{L}=0.9 \mathcal{L}_{\text{diff}}+\lambda\mathcal{L}_{\text{feat}}$; and (ii) \emph{staged} - fine\-tune SVD for one epoch using only $\mathcal{L}_{\text{diff}}$, then enable $\mathcal{L}_{\text{feat}}$ in the second epoch. As shown in Table~\ref{tab:unet_lora_abl_study}, the staged schedule consistently outperforms the joint schedule. Introducing the feature loss immediately underperforms, whereas allowing the model to first adapt its diffusion dynamics to the target dataset yields stable gains once $\mathcal{L}_{\text{feat}}$ is activated. We therefore adopt the staged schedule in all of our experiments.

Next, we analyze whether it is better to fine-tune the U-Net backbone or to train LoRA adapters \cite{lora}, when adding our feature loss to SVD training pipeline. We compare two settings: (i) full U-Net fine-tuning, and (ii) LoRA adapters with rank $r=16$ training. As summarized in Table \ref{tab:unet_lora_abl_study}, across all metrics, fine-tuning the U-Net consistently performs better. This observation aligns with prior works that fine-tune the backbone directly for improving temporal consistency of video generation \cite{sg_i2v, vista}. In contrast, LoRA is typically used for parameter-efficient adaptation to new conditioning signals intended for inference-time control rather than to learn richer dynamics \cite{lora, vista}. Consequently, we adopt full U-Net fine-tuning in our experiments.

\vspace{1em}

\noindent\textbf{Detector-agnostic detection gains.}
We evaluate detection performance with two off-the-shelf detectors - DN-DETR \cite{dn_detr} and YOLOv11 \cite{yolov11}, under identical protocols. Across both models, our method consistently improves mAP and mIoU over the baseline, indicating that the gains are \emph{detector-agnostic}. With DN-DETR, mAP increases from 16.75 to 24.92 (+48.8\% rel. gain), and with YOLOv11 from 17.61 to 21.31 (+21.0\% rel. gain). We attribute the larger relative gain and stronger absolute performance of DN-DETR to its denoising-based training, which is more resilient to residual diffusion artifacts. Because of that, we adopt DN-DETR as the primary detector in the main tables and use YOLOv11 as a robustness check.

\newcolumntype{Y}{>{\centering\arraybackslash}X}

\begin{table}[t]
\centering
\caption{Detector-agnostic improvements under the proposed semantic feature consistency loss: DN-DETR vs. YOLOv11.}
\label{tab:detr_yolo_abl_study}
\footnotesize
\setlength{\tabcolsep}{4pt}
\begin{tabularx}{\columnwidth}{@{} l *{4}{Y} @{}}
\toprule
\multirow{2}{*}{\textbf{Model}} &
\multicolumn{2}{c}{\textbf{DN-DETR} \cite{dn_detr}} &
\multicolumn{2}{c}{\textbf{YOLOv11} \cite{yolov11}} \\
\cmidrule(lr){2-3}\cmidrule(lr){4-5}
& \textbf{mAP} $\uparrow$ & \textbf{mIoU} $\uparrow$
& \textbf{mAP} $\uparrow$ & \textbf{mIoU} $\uparrow$ \\
\midrule
SVD Fine-tuned \cite{svd} & 16.75 & 79.08 & 17.61 & 78.60 \\
\rowcolor{gray!10}
SVD + LSA (Ours)          & \textbf{24.92} & \textbf{80.65} & \underline{21.31} & \underline{80.13} \\
\bottomrule
\end{tabularx}
\end{table}

\section{Conclusion}

We presented Localized Semantic Alignment (LSA), a framework for enhancing temporal consistency in diffusion-based video generation for autonomous driving without relying on control signals at inference time. LSA fine-tunes pre-trained models by aligning semantic features between ground-truth and generated clips through a localized semantic feature consistency loss around dynamic objects, combined with the standard diffusion loss. Despite requiring only a single epoch of fine-tuning, LSA improves temporal coherence and visual fidelity across nuScenes and KITTI datasets. To further assess temporal consistency, we adapted object-centric metrics, mAP and mIoU, demonstrating the robustness of our approach. LSA offers an efficient, generalizable method for producing temporally coherent driving videos, paving the way toward scalable video generation frameworks for autonomous systems.

\textbf{Limitations and future work.} 
While our method effectively improves temporal consistency in diffusion-based video generation for driving scenarios, it relies on ground-truth bounding boxes during fine-tuning. Future work could replace these annotations with pseudo-labels from off-the-shelf object detection models, enabling large-scale fine-tuning on unlabeled data. Additionally, since our loss operates at the feature level, it could be applied to other generative architectures to further study the role of semantic alignment in promoting temporal consistency. Future work should also explore generating multiple plausible futures and establishing evaluation metrics that meaningfully assess temporal consistency of these predictions. 





\end{document}